\documentclass[final]{cvpr}

\usepackage{times}
\usepackage[pdftex]{graphicx}
\usepackage{amsmath}
\usepackage{amssymb}
\usepackage{cite}

\usepackage{color,colortbl}
\usepackage{multirow}
\usepackage{tabularx}
\usepackage{color}
\usepackage[table]{xcolor}
\usepackage{overpic}
\usepackage{wrapfig,lipsum,booktabs,subfigure}
\usepackage{url}
\usepackage{makecell}

\usepackage{algorithm}
\usepackage{algpseudocode}
\usepackage{amsmath}
\usepackage{graphics}

\graphicspath{{./fig/}}

\usepackage[pagebackref=true,breaklinks=true,colorlinks,bookmarks=false]{hyperref}

\def\cvprPaperID{1360} 
\def\confYear{CVPR 2021}
\renewcommand{\eqref}[1]{Eqn.~(\ref{#1})}
\renewcommand{\algref}[1]{Algorithm~(\ref{#1})}

\newcommand{\gsh}[1]{{\textcolor{black}{#1}}}

\newcommand{\myPara}[1]{\vspace{-.12in}\paragraph{#1}}
\newcommand{\sArt}{{state-of-the-art~}}
\newcommand{\tbf}[1]{\textbf{#1}}
\newcommand*\samethanks[1][\value{footnote}]{\footnotemark[#1]}
\def\ie{\emph{i.e.,~}}

\def\etc{\emph{etc}}
\def\etal{{\em et al.}}

\newcommand{\ConfInf}{\vspace{-.7in} {\normalsize \normalfont \color{blue}{
	IEEE International Conference on Computer Vision and Pattern Recognition (CVPR 2021)}} \vspace{.45in} \\}

\begin{document}


\title{\ConfInf Global2Local: Efficient Structure Search for Video Action Segmentation}

\author{Shang-Hua Gao$^1$\thanks{Equal contribution} \qquad Qi Han$^1$\samethanks \qquad Zhong-Yu Li$^1$ \\
  \qquad Pai Peng$^2$ \qquad Liang Wang$^3$ \qquad Ming-Ming Cheng$^1$ \thanks{M.M. Cheng (cmm@nankai.edu.cn) is the corresponding author.} \\
  TKLNDST, CS, Nankai University$^1$ \qquad  Tencent$^2$ \qquad NLPR$^3$\\
  {\tt\small http://mmcheng.net/g2lsearch}
}

\maketitle
\pagestyle{empty}  
\thispagestyle{empty} 
\begin{abstract}
Temporal receptive fields of models play an important role in action segmentation.
Large receptive fields facilitate the long-term relations among video clips
while small receptive fields help capture the local details. 
Existing methods construct models with hand-designed receptive fields in layers.
Can we effectively search for receptive field combinations
to replace hand-designed patterns?
To answer this question, 
we propose to find better receptive field combinations 
through a global-to-local search scheme.
Our search scheme exploits both global search to find the coarse combinations 
and local search to
get the refined receptive field combination patterns further.
The global search finds possible coarse combinations 
other than human-designed patterns.
On top of the global search, 
we propose an expectation guided iterative local search scheme to 
refine combinations effectively. 
Our global-to-local search can be plugged into existing action segmentation
methods to achieve \sArt performance.
The source code is publicly available on \url{https://github.com/ShangHua-Gao/G2L-search}.
\end{abstract}


\section{Introduction}

Action recognition segments the action of each video frame,
playing an important role in computer vision applications 
such as clips tagging~\cite{soleymani2011multimodal}, 
video surveillance~\cite{collins2000introduction, collins2000system}, 
and anomaly detection~\cite{saligrama2012video}.
While conventional works~\cite{simonyan2014two,feichtenhofer2017spatiotemporal,carreira2017quo,feichtenhofer2019slowfast} 
have continuously refresh the recognition performance of short trimmed videos
containing a single activity,
segmenting each frame densely in long untrimmed videos remains
challenging as those videos contain many activities 
with different temporal lengths.
Temporal convolutional networks (TCN)
\cite{lea2017temporal,farha2019ms,MS-TCN-PAMI20,wangboundary,fayyaz2020sct} 
are widely adapted in action segmentation tasks with their ability
to capture both long-term and short-term information.
Appropriate receptive fields in layers are crucial for TCN as 
large receptive fields contribute to long-term dependencies 
while small receptive fields benefit the local details.
State-of-the-art (SOTA) methods
\cite{MS-TCN-PAMI20,chen2020action,wangboundary,li2020set,huang2020improving} 
rely on human-designed receptive field combinations, 
\ie dilation rate or pooling size in each layer, 
to make the trade-off between capturing long and short term dependencies.
Questions have raised: Are there other effective receptive field combinations 
that perform comparable or better than hand-designed patterns? 
Will the receptive field combinations vary among different datasets? 
To answer those questions, 
we propose to find the possible receptive field combinations in a 
coarse-to-fine scheme through the global-to-local search.

\begin{figure}[!t]
   \centering
   \begin{overpic}[width=\linewidth]{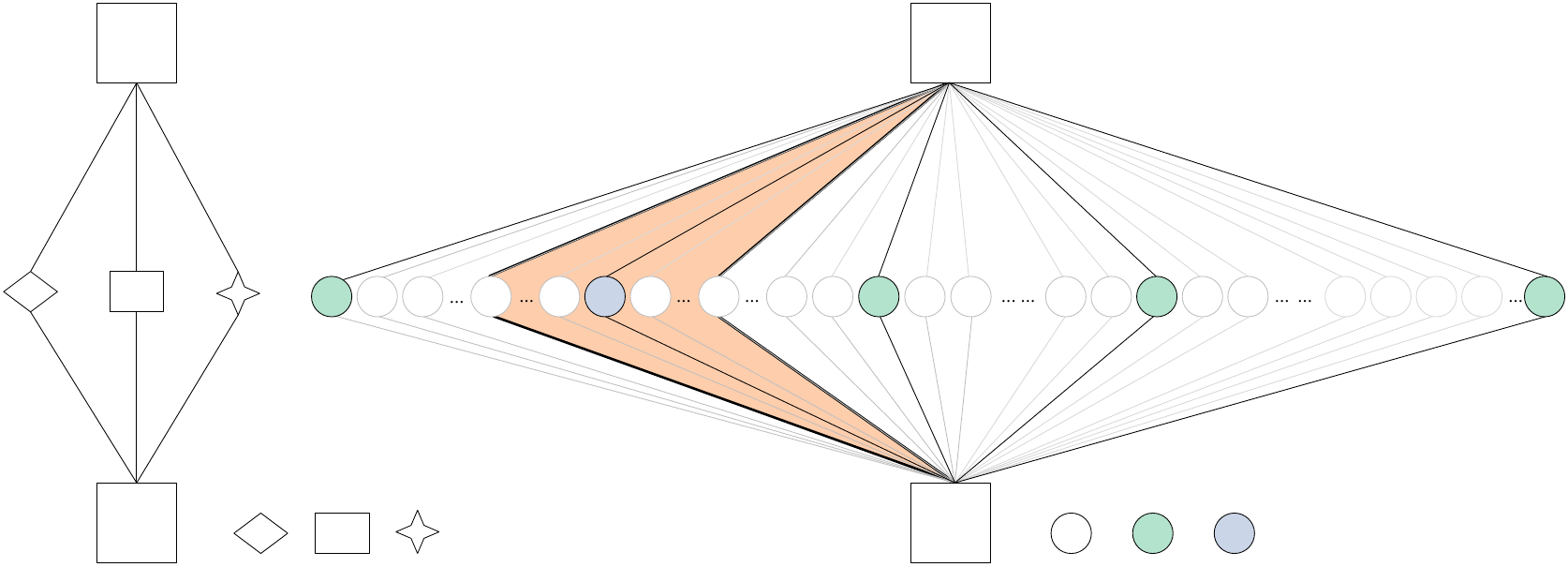}
        \put(30, 1){operators}
        \put(83, 1){dilations}
   \end{overpic}
   \caption{Search space comparison between searching for network architecture 
      and receptive field combinations. 
      Left: Network architecture search mostly search for several operations 
      with different functions. 
      Right: The search space of receptive field combinations is huge. 
      The white, green, blue nodes and orange shade represent 
      the dilation rate candidates, 
      the sparse search space in global search, 
      one of the global searched results, 
      and the local search space.
   }\label{fig:tasks}
\end{figure}

As shown in~\figref{fig:tasks},
unlike the existing network architecture search spaces
\cite{liu2019darts,cai2018proxylessnas,howard2019searching} 
that only contain several operation options within a layer,
the available search space of receptive field combinations could be huge. 
Suppose a TCN has $L$ convolutional layers and $D$ possible receptive fields 
in each layer.
There are $D^L$ possible combinations,
\ie the number of possible receptive field combinations 
in MS-TCN~\cite{farha2019ms} is $1024^{40}$.
Directly apply network architecture searching algorithms
\cite{howard2019searching,liu2019auto,liu2019darts,xie2017genetic} 
to such a huge search space is impractical.
For example, conventional reward-based searching methods
\cite{real2019regularized,liu2017hierarchical,xie2017genetic} 
are not suitable for CNN-based models with a huge search space.
The model training and performance evaluation of each possible combination are too costly.
Differentiable architecture searching methods (DARTS)
\cite{liu2019darts,cai2018proxylessnas,liu2019auto} 
rely on shared big networks to save training time,
thus only supporting several operators within a layer due to 
the model size constraint.  
Moreover, they heavily dependent on the initial combination and fail to 
find new combinations with a huge difference from the initial one.
While our goal is to explore effective receptive field combinations other than
human-designed patterns in the huge search space, 
those algorithms are either too costly or cannot support the large search space.

To explore the search space with low cost,
we exploit both a genetic-based global search to find 
the coarse receptive field combinations 
and an expectation guided iterative (EGI) local search to get 
the refined combinations.
Specifically, we follow the MS-TCN~\cite{farha2019ms}
to use dilation rates to determine layers' receptive fields.  
A genetic-based global search scheme is proposed to find coarse combinations 
within a sparsely sampled search space at an affordable cost.
The global search discovers various combinations that achieve 
even better performance
than human designings but have completely different patterns.
Based on the global-searched coarse combinations,
we propose the local search to determine fine-grained dilation rates.
Our proposed convolutional weight-sharing scheme enforces learned dilation weights to approximate the probability mass distribution for calculating the expectation of dilation rates.
The expectation guided searching transfer the discrete 
dilation rates into a distribution,
allowing fine-grained dilation rates searching.
With an iteratively searching process,
the local search gradually finds more effective fine-grained receptive field
combinations with low cost.
Our proposed global-to-local search scheme can be plugged into existing models, 
surpassing human-designed structures with impressive performance gain.
In summary, we make two major contributions:
\begin{itemize}
  \item The expectation guided iterative local search scheme enables 
    searching fine-grained receptive field combinations 
    in the dense search space.
  \item The global-to-local search discovers effective receptive field
    combinations with better performance than hand-designed patterns.
\end{itemize}

\section{Related Work}

\subsection{Action Segmentation}
Many approaches have been proposed for modeling dependencies for 
action segmentation.
Early works~\cite{fathi2013modeling,fathi2011understanding,fathi2011learning} 
mostly model the changing state of appearance and actions with sliding 
windows~\cite{rohrbach2012database,karaman2014fast,bhattacharya2014recognition}.
Thus they mainly focus on short-term dependencies.
Capturing both short-term and long-term dependencies then gradually becomes 
the focus of action segmentation.

\myPara{Sequential Model.}
Sequential models capture long-short term dependencies in an iterative form.
Vo and Bobick \cite{vo2014stochastic} apply the Bayes network to 
segment actions represented with the stochastic context-free grammar.
Tang \etal~\cite{tang2012learning} use a hidden Markov model
to model transitions between states and durations.
Later, hidden Markov models are combined with 
context-free grammar~\cite{kuehne2016end}, 
Gaussian mixture model~\cite{kuehne2017weakly}, 
and recurrent networks~\cite{richard2017weakly,kuehne2018hybrid} 
to model long-term action dependencies.
Cheng~\etal~\cite{cheng2014temporal} apply the sequence memorizer 
to capture long-range dependencies in visual words learned from the video.
However, these sequential models are inflexible in parallel modeling 
long-term dependencies and usually suffer from 
information forgetting \cite{farha2019ms,MS-TCN-PAMI20}.

\myPara{Multi-stream Architecture.}
Some researchers
\cite{richard2016temporal,singh2016multi,singh2016first,ding2017tricornet} 
utilize multi-stream models to model dependencies 
from both the long and short term.
Richard and Gall employ \cite{richard2016temporal} dynamic programming 
to inference models composed of length model, 
language model, and action classifier.
Singh~\etal~\cite{singh2016multi} learn short video chunks representation 
with a two-stream network and pass these chunks to a bi-directional network 
to predict action segmentation results sequentially.
A three-stream architecture is proposed in \cite{singh2016first}, 
which contains egocentric cues, spatial and temporal streams.
Tricornet~\cite{ding2017tricornet} utilizes a hybrid temporal convolutional 
and recurrent network to capture local motion and 
memorize long-term action dependencies. 
CoupledGAN~\cite{gammulle2019coupled} uses a GAN model to utilize 
multi-modal data to better model human actions' evolution.
Capturing long-short term information with multiple streams increases 
the computational redundancy.

\myPara{Temporal Convolutional Network.}
Recently, temporal convolutional networks (TCN) are introduced to model 
dependencies of different ranges within a unified structure by adjusting receptive fields and can process long videos in parallel.
Lea~\etal~\cite{lea2017temporal} propose the encoder-decoder style TCN 
for action segmentation to capture long-range temporal patterns 
and apply the dilated convolution to enlarge the receptive field.
TDRN~\cite{lei2018temporal} further introduces the deformable convolution 
to process the full-resolution residual stream and low-resolution pooled stream.
MS-TCN~\cite{farha2019ms,MS-TCN-PAMI20} utilizes multi-stage dilated TCNs with 
hand-designed dilation rate combinations to capture information from 
various temporal receptive fields.
However, the adjustment of receptive fields still relies on human design,
which may not be appropriate.
Our proposed efficient receptive field combinations searching scheme 
can automatically discover more efficient structures, 
improving these TCN based methods.

\myPara{Complementary Techniques.}
Instead of capturing long-term and short-term information, 
some works~\cite{ding2018weakly,wangboundary} further improve the 
action segmentation performance with boundary refinement. 
Li~\etal~\cite{ding2018weakly} utilize an iterative training procedure 
with transcript refinement and soft boundary assignment.
Wang~\etal~\cite{wangboundary} leverage semantic boundary information 
to refine the prediction results.
Other researchers focus on action segmentation under the 
weakly supervised~\cite{kuehne2017weakly,richard2017weakly,ding2018weakly} 
or unsupervised ~\cite{sener2018unsupervised} settings.
These works still rely on the efficient TCN to model the action dependencies, 
thus complementing the proposed method.

\subsection{Network Architecture Search}

The genetic algorithm~\cite{mitchell1998introduction} has achieved remarkable 
performance on a wide range of applications.
Many genetic-based methods are recently introduced for the 
neural networks architecture search of vision tasks~\cite{real2019regularized,liu2017hierarchical,xie2017genetic,sun2020automatically,lu2019nsga}.
An evolutionary coding scheme is proposed in Genetic CNN~\cite{xie2017genetic}
to encode the network architecture to a binary string. 
A hierarchical representation is presented by Liu~\etal~\cite{liu2017hierarchical} 
to constrain the search space.
Real~\etal~\cite{real2019regularized} regularize the evolution by an 
age property selection operation.
Sun~\etal~\cite{sun2020automatically} introduce a variable-length 
encoding method for effective architecture designing.
However, the genetic algorithm requires the training of each candidate, 
consuming too much computational cost when faced with a huge search space.

Differentiable architecture search~\cite{liu2019darts} saves the training time 
by introducing a large network containing subnetworks 
with different searching options. 
The importance of searched blocks is 
determined by gradient backpropagation~\cite{rumelhart1986learning}.
This differentiable search idea is further extended \cite{zela2020understanding}
to deal with semantic segmentation \cite{liu2019auto} and 
other tasks  beyond image classification~\cite{cai2018proxylessnas}.
However, these network architecture search methods are designed for finding 
a limited number of operations such as convolution, ReLU, 
batch normalization, short connection,~\etc.
Thus, they cannot handle the huge receptive field combinations 
search space.
In this paper, 
we propose a global search to handle the huge search space with sparse sampling.
The expectation guided iterative local search then transfers the sparse search space of 
receptive fields into the dense one for fine-level searching.

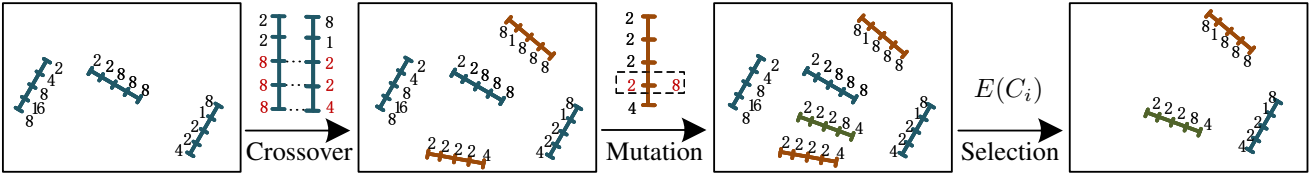
\begin{figure*}[t!]
   \centering
   \begin{overpic}[width=\linewidth]{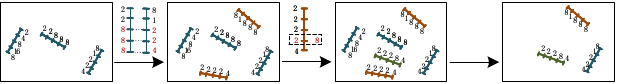}
      \put(18.6, 1.2){Crossover}
      \put(46, 1.2){Mutation}
      \put(73, 1.2){Selection}
      \put(74, 6){$E(C_i)$}
   \end{overpic}
   \caption{Illustration of one iteration in our genetic-based global search algorithm.}
   \label{fig:population}
\end{figure*}

\section{Method}

The pipeline of our proposed global-to-local search method has two components: 
(i) a genetic-based global search algorithm that produces coarse but 
competitive combinations of the receptive fields;
(ii) an expectation guided iterative local search scheme that locally refines the global-searched coarse structures.

\subsection{Description} \label{sec:de}
Our objective is to efficiently search for optimal receptive field combinations 
for the given dataset. 
The receptive field can be represented with multiple forms, 
such as the dilation rate, kernel size, pooling size, stride, 
and the stack number of layers.
In this work, we mainly follow the MS-TCN~\cite{farha2019ms} to formulate
the receptive fields using the combinations of dilation
rates in layers and propose to evolve these combinations during the searching process.
Note that other receptive field representations can also be applied to 
the proposed global-to-local search with some minor adjustments.

Suppose a TCN has $L$ convolutional layers and $D=\{d_1,d_2,...,d_N\}$ 
is the possible dilation-rates/receptive-fields in each layer.
The combination of receptive fields is represented with 
$C=\{c_1,...,c_l,...,c_L\}$,
where $l\in [1, L]$ is the index of layers with dilated convolutions,
and $c_l\in D$ is the receptive field of each layer.
There are $|D|^L$ possible combinations of receptive fields,
\ie the possible receptive field combinations in MS-TCN~\cite{farha2019ms} is $1024^{40}$ when dilation rates ranging from 1 to 1024.
Directly searching for effective combinations in such a large search space is impractical.
We thus decompose the searching process into the global and local search to
find the combination in a coarse-to-fine manner.
\subsection{Global Search} \label{sec:gs}

The objective of the global search is to find the coarse receptive field combinations
with affordable cost.
Therefore, we reduce the search space by sparsely sampling the dilation rates within layers.
Multiple sparse discrete sampling strategies such as uniform sampling, 
gradually sparse sampling, and gradually dense sampling can be applied to sparse the search space.
A gradually sparse sampling scheme from small to large dilation rates is appropriate for the action segmentation task.
Because small receptive fields benefit the extraction of precise local details
while large receptive fields contribute to coarse long-term dependencies of video sequences.
Therefore we formulate the receptive field space in global search as:
\begin{equation}
    D_g=\{d_i = k^{i}, i \in [0,1,\cdots T] \}, 
   \label{eq:d_g_space}
\end{equation}
where $k$ is the controller of the search space sparsity, and $T$ determines the
largest available receptive field.
With the same maximum receptive field, $|D_g| \ll |D|$. The search space is greatly reduced.
\ie when set $k=2$, and set the maximum receptive field to 1024 as in MS-TCN, 
the search space is reduced from $1024^{40}$ to $11^{40}$.

However, the reduced space of receptive field combinations can still be huge, 
unaffordable for a brute force search.
We propose a genetic algorithm~\cite{mitchell1998introduction} based method 
to find coarse combinations that are competitive or even better 
than human designing.
We illustrate one iteration of our proposed genetic-based global search in
\figref{fig:population}.
We now detail the selection, crossover, mutation process within 
our proposed global search method.

\myPara{Selection.} 
The population of receptive field combinations can be described as 
a group of candidate structures $P=\{C_{i}, i \in [1, M]\}$, 
where $C_{i}$ is the candidate structure in the global search space, 
and $M$ is the number of individuals in the population. 
The selection operation selects individuals to be kept in $P$ based on the 
estimated performance of each structure $C_{i}$, denoted by $E(C_{i})$:
\begin{equation}\label{eq:e_c}
   E(C_{i}) = f(V|C_{i}, \theta_n),   
\end{equation}
where $f(\cdot)$ is the evaluation metric detailed in \secref{sec:experimets}, 
and $V$, $\theta_n$ are 
the cross-validation set and model trained with $n$ epochs, respectively.
\myPara{Crossover.} 
This operation generates new samples of receptive field combinations. 
Every two combinations in the population are exchanged to born new patterns of 
the combination while maintaining the local structures.
Each $C_{i}$ will be selected for the crossover operation with probability $p(C_{i})$: 
\begin{equation}\label{eq:p_c}
   p(C_{i})= \frac{E(C_{i})}{\sum_{i}^{M}E(C_{i})}.
\end{equation}
Instead of randomly exchanging individual points, 
we choose to exchange random segments of the receptive field combination 
since the representation ability lies in the combination patterns.
Specifically, we randomly choose two anchors and exchange combinations within anchors to generate new samples. 

\myPara{Mutation.} 
The mutation operation avoids getting stuck 
in local optimal results by choosing an 
individual with probability $p_m$ and 
randomly changing a value within the selected combination.

\begin{algorithm}[htb] 
  \caption{Global Search.} 
  \label{alg:global_search} 
  \begin{algorithmic}
    \renewcommand{\algorithmicrequire}{\tbf{Input:}}
    \Require Iterations $N$, training epoch $n$, randomly initialized $P$, 
    mutation probability $p_m$, and population size $M$;
    \For{iter in $[1, N]$}
      \State Select individuals for crossover based on \eqref{eq:p_c} and 
      crossover for every two selected individuals;
      \State Mutate the new individuals with probability $p_m$;
      \State Training each individual with $n$ epochs;
      \State Select the top $M$ individuals based on \eqref{eq:e_c} 
      as the new population $P$;
   \EndFor \\
   \Return $P$.
  \end{algorithmic} 
\end{algorithm}

The global search process can be summarised as \algref{alg:global_search},
and a simple example is given in~\figref{fig:population}. 
With the coarse search space and the global search method,
we can find receptive field combinations with different patterns than 
human-designed structures while having similar or even better performance.
We further propose the local search to locally find the more efficient 
combinations on top of the global-searched structures.
We show in \tabref{tab:ablation_local_init} that local search heavily relies 
on the initial structure,
revealing the importance of global search.


\subsection{Expectation Guided Iterative Local Search} \label{sec:ls}

The local search aims to find more efficient receptive field combinations 
in a fine-grained level at a low cost.
A naive approach is to sample finer-grained dilation rates near the 
initial dilation rate searched by the global search and 
apply existing DARTS algorithms~\cite{cai2018proxylessnas,liu2019darts} 
to choose for the proper one.
However, even with the good initial structure provided by the global search,
the available range of fine-grained dilation rates is still large.
Existing search algorithms are designed for searching sparse operators 
with several choices in each layer,
thus cannot handle dilation rates with hundreds of choices.
While too sparsely sampling is in conflict with our goal of searching 
for the finer-grained receptive fields.
\gsh{
Also, DARTS methods search operators with different functionality~\cite{liu2019darts},
while the searching on receptive fields only contains one functional dimension.
Different subsets in the dataset sometimes prefer different searching options. 
Searching within a functional dimension enables us to 
determine dilation rates with the expectation of all subsets
instead of choosing the option required by one majority subset.
Therefore, we propose an expectation guided iterative (EGI) 
local search scheme to determine the finer-level dilation rates 
on top of the global-searched structures.
}

\gsh{
Suppose that the receptive field of a layer $l$ is $D_{l}$.
For a dataset, once we get the probability mass distribution of dilation rates around $D_{l}$,
we can obtain the expected dilation rate with the weighted average 
of the dilation rates required by all subsets.
However, the probability mass of dilation rates for the dataset is inaccessible.
Therefore, we utilize a convolutional weight-sharing scheme to enforce
the learned importance weights of dilation rates to approximate the probability mass.
To get the approximated probability mass function of dilation rates,} 
we first evenly sample $S$ dilation rates near the initial dilation rate 
$D_{l}$ within the range of $[D_l \pm \Delta D_{l}]$.
The set of available dilation rates within this layer is
$T_l = \{ d_i | i \in [1, S] \}$, 
where $d_i = D_{l}-\Delta D_{l} + (i-1) \cdot 2 \Delta D_{l}/(S-1)$.
$\Delta D_{l}$ is the finer controller of the search space that is smaller 
than the sampling sparsity in the global search.

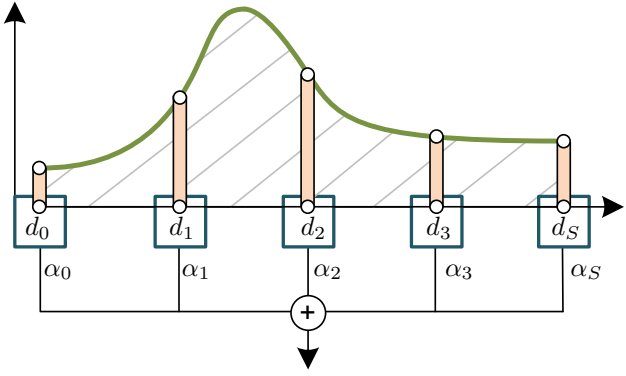
\begin{figure}[t]
   \centering
   \begin{overpic}[width=\linewidth]{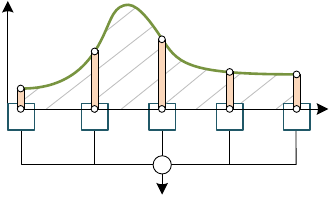}
      \put(4, 22.5){$d_{0}$}
      \put(27, 22.5){$d_{1}$}
      \put(48, 22.5){$d_{2}$}
      \put(68, 22.5){$d_{3}$}
      \put(88, 22.5){$d_{S}$}
      \put(7, 16){$\alpha_{0}$}
      \put(29, 16){$\alpha_{1}$}
      \put(50, 16){$\alpha_{2}$}
      \put(71, 16){$\alpha_{3}$}
      \put(91, 16){$\alpha_{S}$}
      \put(47.9, 9.1){\tbf{+}}
   \end{overpic}
   \caption{\gsh{The approximated probability mass function of dilation rates is determined by the multi-dilated 
      convolutional layer with shared weights. $d_i$ is the dilation rate and $\alpha_i$ is the PMF in~\eqref{equ:normalize}.}
   } 
   \vspace{-10pt}
   \label{fig:local_conv}
\end{figure}

\begin{algorithm}[htb] 
   \caption{Expectation Guided Iterative Local Search.} 
   \label{alg:local_search} 
   \renewcommand{\algorithmicrequire}{\tbf{Input:}}
   \begin{algorithmic}
     \Require Iterations $N$, initial receptive fields $D$;
     \State Initialize model using given $D$;
     \For{iter in $[1, N]$}
       \State Construct $T_l$ for each layer based on $D$;
       \State Train model to get the $PMF$ in ~\eqref{equ:normalize};
       \State Obtain new dilation rates through \eqref{equ:recombined};
       \State Update $D$;
     \EndFor \\
     \Return local-searched $D$.
   \end{algorithmic} 
\end{algorithm}

\begin{figure}[t!]
   \centering
   \vspace{8pt}
   \begin{overpic}[width=\linewidth]{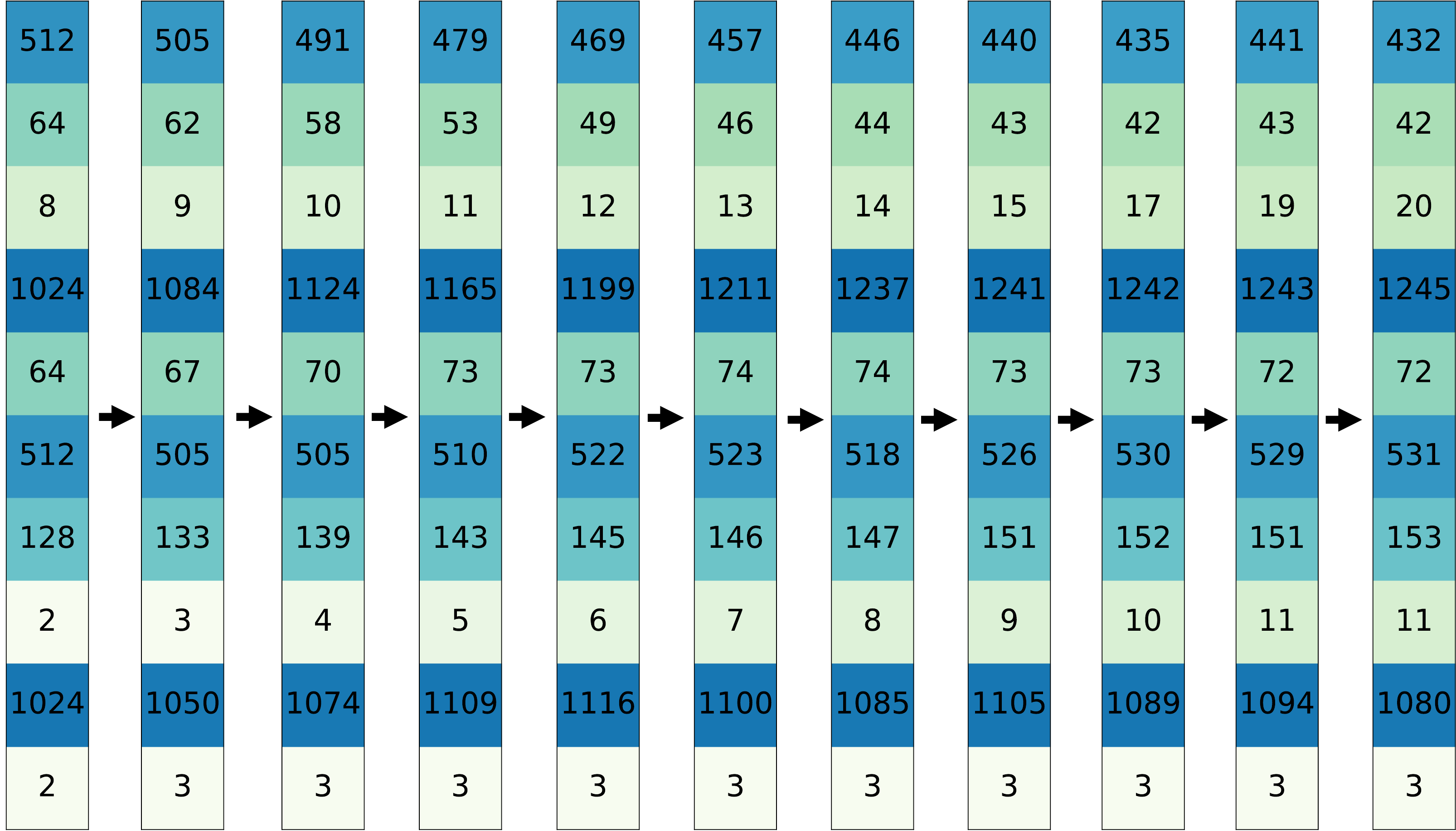}
      \put(46.5, 59){Output}
      \put(47, -4.5){Input}
   \end{overpic}
   \vspace{2pt}
   \caption{Visualization of receptive field combinations changes during the EGI  local searching process.} 
   \label{fig:dilation_changes}
\end{figure}

With the dilation rates set $T_l$, 
we propose a multi-dilated layer composed of a shared 
convolutional weight and multiple branches with different dilation rates, 
as shown in~\figref{fig:local_conv}.
Each branch has a unique weight to determine the importance of the 
dilation rate.
During the searching process, 
the weights are updated with the gradient backpropagation \gsh{to reflect the receptive field requirements of the dataset}.
Existing DARTS schemes~\cite{liu2019darts,zela2020understanding} have separated weights in each branch. In contrast, our convolutional weight-sharing strategy forces the model to learn the \gsh{approximated} probability of receptive fields and ease the model convergence.
Specifically, the dilation rates in the multi-dilated convolutional layer 
are set to $T_l$.
Apart from the shared convolutional $\theta$,  
the multi-dilated layer contains weight 
$W = \{w_1, w_2, ..., w_i, i \in [1, S]\}$ 
to determine the importance of the dilation rates.
$W$ is unbounded, 
thus cannot be directly used to determine the dilation rates probability.
Therefore, we propose a normalization function to get the \gsh{approximated probability} mass 
function $PMF(d_i)$ of dilation rates through normalizing $w_i$:
\begin{equation}\label{equ:normalize}
  PMF(d_i) = \alpha_i = \frac{|w_i|}{\sum_{i}^{S} |w_i| }. 
\end{equation}
With the probability mass function, given the input feature $x$,
the output $y$ of the multi-dilated convolutional layer can be written 
as follows:
\begin{equation}\label{equ:forward}
  y = \sum_{i}^{S} \alpha_i \Psi(x, d_i, \theta), 
\end{equation}
where $\Psi(x, d_i, \theta)$ is the convolutional operation 
with the shared weight $\theta$ and dilation rate $d_i$.
$\alpha_i$ is updated with gradient optimization.
Once we get the probability mass function,
the newly searched dilation rate $D_l^{'}$ is obtained with the expectation:
\begin{equation}\label{equ:recombined}
   D_l^{'} = \lfloor \sum_{d_{i}\in T_l} PMF(d_i) \cdot d_{i} \rfloor. 
\end{equation}
To reduce the computational cost during the local search process, 
we reduce the number of dilation rates in $T_l$ to 3 by default
and apply the iterative search scheme to find the more suitable dilation rate 
based on the  $D_l^{'}$ from the last iteration.
The local search process can be summarised as \algref{alg:local_search}.
Furthermore,~\figref{fig:dilation_changes} visualizes the dilation rates 
changes during the local searching process.

\def\fMeaures{\multicolumn{3}{c}{F@\{10,25,50\}}}

\begin{table*}[t]
  \small
  \centering 
  \setlength{\tabcolsep}{2.23mm}
  \begin{tabular}{lccccc|ccccc|ccccc}
    \toprule & \multicolumn{5}{c|}{\tbf{BreakFast}} &  
    \multicolumn{5}{c|}{\tbf{50Salads}} &
    \multicolumn{5}{c}{\tbf{GTEA}} 
    \\
    & \fMeaures & Edit & Acc & \fMeaures & Edit & Acc & \fMeaures & Edit & Acc
    \\ \Xhline{0.8pt}
    \textsl{MS-TCN~\cite{farha2019ms}}
    & 52.6 & 48.1 & 37.9 & 61.7 & 66.3
    & 76.3 & 74.0 & 64.5 & 67.9 & 80.7
    & 87.5 & 85.4 & 74.6 & 81.4 & 79.2
    \\
    \textsl{Reproduce}
    & 69.1 & 63.7 & 50.1 & 69.9 & 67.3
    & 78.8 & 75.3 & 64.4 & 71.4 & 77.8
    & 87.1 & 83.6 & 70.4 & 81.1 & 75.5
    \\ \hline
    \textsl{Global}
    & 72.2 & 66.0 & 51.5 & 71.0 & 69.2
    & 79.3 & 76.5 & 68.1 & 71.9 & 81.2
    & 89.1 & 87.1 & 74.4 & 84.2 & \tbf{78.6}
    \\
    \textsl{Local}
    & \tbf{74.9} & \tbf{69.0} & \tbf{55.2} & \tbf{73.3} & \tbf{70.7}
    & \tbf{80.3} & \tbf{78.0} & \tbf{69.8} & \tbf{73.4} & \tbf{82.2}
    & \tbf{89.9} & \tbf{87.3} & \tbf{75.8} & \tbf{84.6} & 78.5 \\
    \bottomrule
  \end{tabular}
  \vspace{4pt}
  \caption{Performance of the global and local searching stages of 
    our global-to-local 
    searching method using MS-TCN~\cite{farha2019ms} as the baseline.  
  }\label{tab:search}
\end{table*}

\def\GTEA{\textsl{GTEA}~\cite{fathi2011learning}}
\def\Salads{\textsl{50Salads}~\cite{stein2013combining}}
\def\BreakFast{\textsl{BreakFast}~\cite{kuehne2014language}}
\begin{table}[htb]
  \small
  \centering 
  \setlength{\tabcolsep}{2mm}
  \begin{tabular}{lcccc}  \toprule
            & \#Cls &\#Vid&\#Frame& Scene  \\ \midrule
    \GTEA      & 11 & 28 & 1115   & daily activities  \\ \hline
    \Salads    & 17 & 50 & 11552  & preparing salads  \\   \hline 
    \BreakFast & 48 & 1712 & 2097 & cooking breakfast \\ \bottomrule 
  \end{tabular}
  \vspace{2pt}
  \caption{Details of three action segmentation datasets. 
     \#Cls and \#Vid are the numbers of classes and videos, respectively.
     \#Frame is the average frames of videos.
  }\label{tab:datasets}
\end{table}

\section{Experiments} \label{sec:experimets}
We introduce the implementation details, verify the effectiveness, and analyze the property of
our proposed global-to-local search scheme in this section.
\subsection{Implementation Details}
\vspace{3pt}
\label{sec:dataset}
\myPara{Structure Searching and Training.}  
Our proposed method is implemented with the PyTorch~\cite{paszke2019pytorch}, MindSpore~\cite{mindspore2020}, and Jittor~\cite{hu2020jittor} frameworks.
Following existing works~\cite{MS-TCN-PAMI20,farha2019ms}, features are first extracted from videos using the I3D network~\cite{carreira2017quo} and then
passed to action segmentation models to get the temporal
segmentation.
Since our proposed global-to-local search scheme is model-agnostic, 
the training settings for model evaluation, 
\ie training epochs, optimizer, learning rate, batch size, keep the same with 
the cooperation methods~\cite{MS-TCN-PAMI20,wangboundary,chen2020action}. %
In the global search stage, we set the total iterations $N = 100$, 
$k=2$ in~\eqref{eq:d_g_space}, the initialized population size $M=50$,
and mutation probability $p_m=0.2$. 
The $T$ in \eqref{eq:d_g_space} is set to 10, indicating the maximum dilation rate of the global search space is 1024. 
We observe that 5 epochs of training can reflect the structure performance,
and therefore models are trained with 5 epochs for evaluation. In the EGI local search stage,
$\Delta D_l$ and $S$ are set to be $0.1D_l$ and 3, respectively. We train the model for 30 epochs during 
local search and update the structure every 3 epochs.

\myPara{Datasets.} 
Following~\cite{farha2019ms,MS-TCN-PAMI20,wangboundary,chen2020action},
we evaluate our proposed method on three popular action segmentation datasets:
Breakfast~\cite{kuehne2014language}, 50Salads~\cite{stein2013combining}, and GTEA~\cite{fathi2011learning}. 
The details of the three datasets are summarised in~\tabref{tab:datasets}.
As far as we know, the Breakfast dataset is the 
largest public dataset for action segmentation task, which has a larger number of categories and samples
compared with the other two datasets.
So we perform our ablations mainly on the Breakfast dataset if not otherwise stated.
Following common settings~\cite{farha2019ms,MS-TCN-PAMI20,wangboundary,chen2020action}, 
we perform 4-fold cross-validation
for the Breakfast and GTEA dataset and 5-fold cross-validation for the 50Salads dataset.

\myPara{Evaluation Metrics.}
We follow previous works~\cite{farha2019ms,MS-TCN-PAMI20,wangboundary,chen2020action} 
to use the frame-wise accuracy (Acc), segmental edit score (Edit)~\cite{lea2017temporal},
and segmental F1 score~\cite{li2015convolutional} at temporal intersection over union with thresholds 0.1, 0.25, 0.5 (F@0.1, F@0.25, F@0.5)
as our evaluation metrics.


\begin{table}[t]
   \small
   \centering 
   \setlength{\tabcolsep}{1.6mm}
   \begin{tabular}{lccccc}  \toprule
   \tbf{BreakFast} 
   & F@0.1 & F@0.25 & F@0.5 & Edit  & Acc  \\
   \midrule
   \textsl{ED-TCN~\cite{lea2017temporal}}
   & - & - & - & - & 43.3  \\
   \textsl{HTK (64)~\cite{kuehne2016end}}
   & - & - & - & - & 52.0  \\
   \textsl{TCFPN~\cite{ding2018weakly}}
   & - & - & - & - & 56.3  \\
   \textsl{GRU~\cite{richard2017weakly}}
   & - & - & - & - & 60.6  \\
   \textsl{GTRM~\cite{huang2020improving}}
   & 57.5 & 54.0 & 43.3 & 58.7 & 65.0  \\
   \hline
   \textsl{MS-TCN~\cite{farha2019ms}}
   & 52.6 & 48.1 & 37.9 & 61.7 & 66.3  \\
   \textsl{Ours-MS-TCN}
    & 74.9 & 69.0 & 55.2 & 73.3 & 70.7  \\
   \hline
   \textsl{MS-TCN++~\cite{MS-TCN-PAMI20}}
   & 64.1 & 58.6 & 45.9 & 65.6 & 67.6 \\
   \textsl{Ours\dag-MS-TCN++}
    & 72.4 & 66.8 & 53.5 & 70.2 & 69.6  \\
   \hline
   \textsl{BCN~\cite{wangboundary}}
    & 68.7 & 65.5 & 55.0 & 66.2 & 70.4 \\
    \textsl{Ours\dag-BCN}
    & 72.5 & 69.9 & 60.2 & 69.0 & 72.9 \\
   \hline
    \textsl{SSTDA~\cite{chen2020action}}
   & 75.0 & 69.1 & 55.2 & 73.7 & 70.2 \\
   \textsl{Ours\ddag-SSTDA}
    & 76.3 & 69.9 & 54.6 & 74.5 & 70.8 \\
   \bottomrule 
   \end{tabular}
   \vspace{2pt}
   \caption{Cooperating with SOTA methods. 
   We perform the whole search pipeline based on MS-TCN~\cite{farha2019ms}. 
   Because of the limited computing resources, 
   we only perform the EGI local search on MS-TCN++~\cite{MS-TCN-PAMI20} and 
   BCN~\cite{wangboundary}, denoted by \dag. SSTDA~\cite{chen2020action} uses MS-TCN~\cite{farha2019ms} as a backbone, 
   so we directly add our searched 
   structure to SSTDA, denoted by \ddag.} 
   \label{tab:compare_sota}
\end{table}

\subsection{Performance Evaluation}
\vspace{3pt}
\myPara{Global2Local Search.}
Our proposed global-to-local search aims to find new combinations of receptive fields better than human designings.
We mainly take MS-TCN~\cite{farha2019ms} as our baseline architecture to perform the global-to-local search.
When testing the MS-TCN on the Breakfast dataset, 
we train all models with the batch size 8 to save training time.
The reproduced results shown in~\tabref{tab:search}
indicates that large batch size achieves much better performance. 
\tabref{tab:search} shows that global-to-local searched structures achieve considerable performance improvements
than human-designed baselines, \ie the searched structure surpasses the reproduced baseline with 5.8$\%$ in terms of F@0.1.
The global-to-local search focuses on the receptive field combinations, 
thus can cooperate with existing SOTA action segmentation methods to further improve their performance.
As shown in \tabref{tab:compare_sota}, on the large scale BreakFast dataset,
global-to-local search consistently boosts the performance of MS-TCN++~\cite{MS-TCN-PAMI20}, BCN~\cite{wangboundary}, and SSTDA~\cite{chen2020action}.

Also, we give comparisons on two small scale datasets, 50Salads and GTEA dataset in \tabref{tab:compare_sota_gtea_50} and supplementary, proving the effectiveness of our proposed global-to-local search.

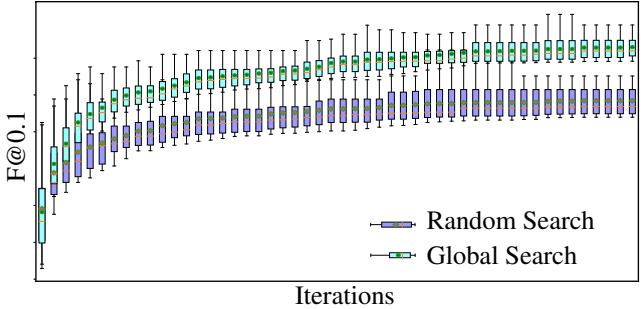
\begin{figure}[!t]
   \centering
   \begin{overpic}[width=\linewidth]{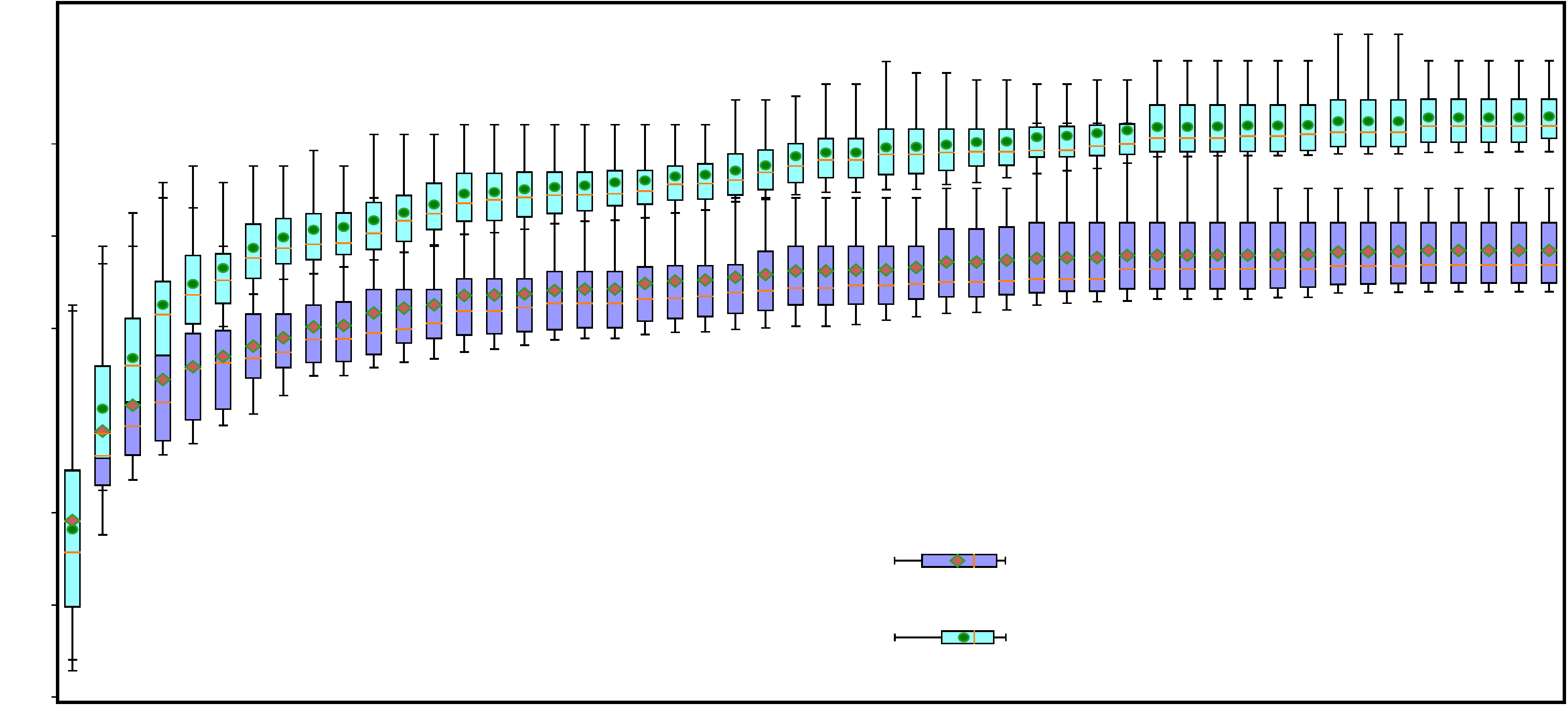}
      \put(66,3){Global Search}
      \put(66,8.5){Random Search}
      \put(-1,16){\rotatebox{90}{F@0.1}}
      \put(45, -3.5){Iterations}
   \end{overpic}
   \vspace{2pt}
   \caption{Performance comparison between our proposed genetic-based search and random search during the global search stage. }
   \label{fig:random_vs_pop}
 \end{figure}

\myPara{Global Search.}
Global search reduces the computational cost with the sparse search space
and our proposed genetic-based searching scheme.
\figref{fig:random_vs_pop} shows the performance change of models during the global searching process.
Compared with the random search, the genetic-based global search convergences faster.
The standard division of model performance searched by genetic-based search is smaller
than the random search, showing the stability of our proposed search scheme. 
The visualized well-performed global-searched structures shown in the supplementary
prove that the global search discovers various structures completely different from
human-designed patterns.
\tabref{tab:ablation_local_init} also shows that the local search heavily relies on global-searched structures
to achieve better performance.

\begin{table}[t]
   \small
   \centering 
   \setlength{\tabcolsep}{2.5mm}
   \begin{tabular}{lccccc}  \toprule
   \tbf{BreakFast} 
   & F@0.1 & F@0.25 & F@0.5 & Edit  & Acc \\
   \midrule
   \textsl{DARTS}
   & 73.8 & 67.6 & 52.8 & 72.8 & 69.3  \\
   \textsl{Ours}
   & \tbf{74.9} & \tbf{69.0} & \tbf{55.2} & \tbf{73.3} & \tbf{70.7} \\
   \bottomrule 
   \end{tabular}
   \vspace{2pt}
   \caption{Performance of our proposed EGI local search and previous DARTS~\cite{liu2019darts}.} 
   \vspace{-10pt}
   \label{tab:ablation_das}
\end{table}

\myPara{Local Search.}
Based on the global-searched structures, our proposed EGI local search aims to fine-tune the receptive field in a 
finer search space. Compared with the DARTS~\cite{liu2019darts} method that only supports several search options, 
the EGI local search iteratively finds the accurate dilations in a dense space,
obtaining structures with better performance, as shown in~\tabref{tab:ablation_das}.
As shown in \tabref{tab:ablation_s},
EGI local search is insensitive to the number of sampling dilation rates $S$,
as it searches dilation rates with the expectation.
\tabref{tab:ablation_local_init} shows that the EGI local search 
can boost the performance of randomly generated, human-designed, and global-searched structures.
Still, the performance of the local-searched structures is related to the initial structures,
as local search focuses on searching for receptive fields within a finer local search space.
We visualize the searching process of the iterative local search in~\figref{fig:dilation_changes}.
The dilation rates for each layer gradually converge to a suitable state during the iterative searching process.
\tabref{tab:ablation_weight_function} verifies different ways to get the approximated probability mass function 
$PMF(d_i)$ from
weight $w$. \eqref{equ:normalize} is more superior than the sigmoid function and softmax function
as it maintains the probability distribution while the other two functions change the distribution non-linearly.

\begin{table}[t]
   \small
   \centering 
   \setlength{\tabcolsep}{2mm}
   \begin{tabular}{lccccc}  \toprule
   \tbf{BreakFast} 
   & F@0.1 & F@0.25 & F@0.5 & Edit  & Acc  \\
   \midrule
   \textsl{random}
   & 67.7 & 61.8 & 48.3 & 68.4 & 67.0  \\
   \textsl{random + local}
   & 73.6 & 67.8 & 53.7 & 72.3 & 69.9  \\
   \hline
   \textsl{baseline~\cite{farha2019ms}}
   & 69.1 & 63.7 & 50.1 & 71.0 & 69.2  \\
   \textsl{baseline + local}
   & 74.1 & 68.5 & 55.3 & 72.3 & 70.2  \\
   \hline
   \textsl{global}
   & 72.2 & 66.0 & 51.8 & 71.5 & 69.4 \\
   \textsl{global + local}
   & 74.9 & 69.0 & 55.2 & 73.3 & 70.7 \\
   \bottomrule 
   \end{tabular}
   \vspace{2pt}
   \caption{Performance of the EGI local search initialized by different structures.} 
   \label{tab:ablation_local_init}
\end{table}

\begin{table}[t]
   \small
   \centering 
   \setlength{\tabcolsep}{2.5mm}
   \begin{tabular}{lccccc}  \toprule
   \tbf{BreakFast} 
   & F@0.1 & F@0.25 & F@0.5 & Edit  & Acc  \\
   \midrule
   \textsl{$S=2$}
   & 74.8 & 68.9 & 55.0 & 73.4 & 70.4  \\
   \textsl{$S=3$}
   & 74.9 & 69.0 & 55.2 & 73.3 & 70.7  \\
   \textsl{$S=4$}
   & 74.9 & 68.8 & 55.1 & 73.3 & 70.9  \\
   \bottomrule 
   \end{tabular}
   \vspace{2pt}
   \caption{Ablation of the value of $S$ in the EGI local search.} 
   \label{tab:ablation_s}
\end{table}

\begin{table}[t]
   \small
   \centering 
   \setlength{\tabcolsep}{2.5mm}
   \begin{tabular}{lccccc}  \toprule
   \tbf{BreakFast} 
   & F@0.1 & F@0.25 & F@0.5 & Edit  & Acc \\
   \midrule
   \textsl{sigmoid}
   & 72.7 & 66.9 & 52.7 & 71.8 & 69.4  \\
   \textsl{softmax}
   & 73.2 & 67.2 & 52.0 & 71.6 & 69.7 \\
   \textsl{\eqref{equ:normalize}}
   & \tbf{74.9} & \tbf{69.0} & \tbf{55.2} & \tbf{73.3} & \tbf{70.7} \\
   \bottomrule 
   \end{tabular}
   \vspace{2pt}
   \caption{Ablation of possible probability mass functions in EGI local search.} 
   \vspace{-8pt}
   \label{tab:ablation_weight_function}
\end{table}

 \begin{figure*}[!t]
   \centering
   \begin{overpic}[width=\linewidth]{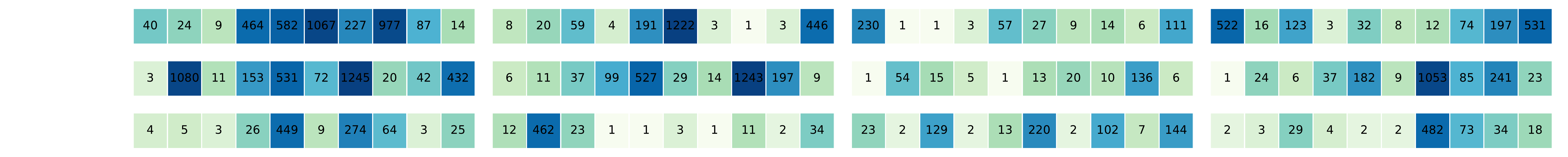}
   \put(0,7.9){50Salads}
   \put(0,4.5){BreakFast}
   \put(0,1){GTEA}
   \end{overpic}
   \caption{Visualization of the global-to-local searched structures of three datasets with the MS-TCN baseline. 
            Each row represents the dilations of one structure, which contains four stages.}
   \label{fig:architectures_datasets}
   \vspace{-8pt}
 \end{figure*}

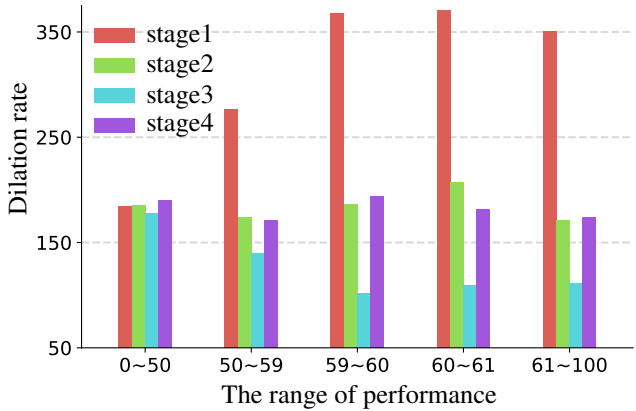
\begin{figure}[!t]
   \centering
   \begin{overpic}[width=\linewidth]{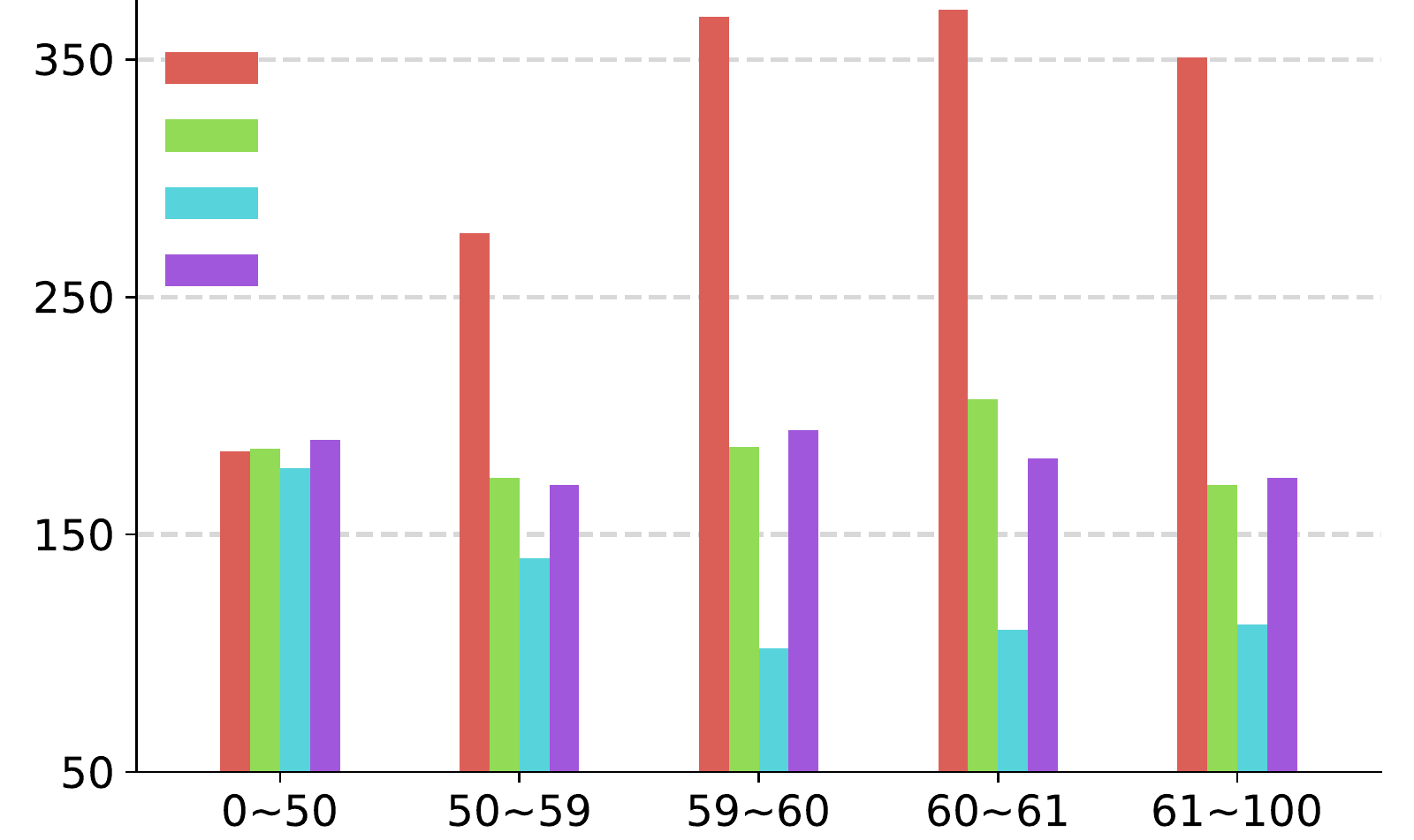}
    \put(20,53.6){stage1}
    \put(20,48.8){stage2}
    \put(20,43.9){stage3}
    \put(20,39.3){stage4}
   
   \put(32,-4){The range of performance}
   \put(-2,26){\rotatebox{90}{Dilation rate}}
   \end{overpic}
   \vspace{2pt}
   \caption{Visualization of average dilation rates in each stage and the range of performance of global-searched structures.} 
   \label{fig:arverage_dilation}
   \vspace{-8pt}
 \end{figure}

\subsection{Observations}
In this section, we try to exploit the common knowledge contained 
in the global-to-local searched structures.

\myPara{Connections between Receptive Fields and Data.}
We want to know if receptive field combinations vary among data.
Therefore, we evaluate the generalization ability of the searched structures 
on the subsets of the same dataset and different datasets, respectively.
Within the BreakFast dataset, we perform the global-to-local search on one fold and then
 evaluate the searched structures on other folds.
\tabref{tab:ablation_cross_val_splits} shows that there is almost no obvious performance gap on different folds, 
indicates that receptive field combinations almost have no difference within a dataset.
However, when search and evaluate structures across different datasets,  
different structures searched on different datasets have a large performance gap as shown in~\tabref{tab:ablation_cross_val_datasets}. 
We can conclude that different data distribution will result in different receptive field combinations.
We visualize the structures searched from different datasets in~\figref{fig:architectures_datasets}.
The structure searched on 50Salads dataset trends to have larger receptive fields, while the structure searched on the GTEA dataset
has smaller receptive fields. 
The number of video frames shown in~\tabref{tab:datasets} is positively correlated with receptive fields. 
Longer videos need larger receptive fields to capture the context.
We also show more searched structures in the supplementary.

\myPara{Receptive Fields for Different Stages.}
Our global-to-local search is based on MS-TCN.
MS-TCN contains four stages, and all stages share the same receptive field combination in human designing.
The visualized searched structures shown in~\figref{fig:architectures_datasets} demonstrate that 
different stages have different receptive field combinations,
which conflicts with human designing. 
We further count the average receptive fields of each stage among all individuals. The range of performance and the average dilation rates of each stage are shown in~\figref{fig:arverage_dilation}.
The average dilation rate in the first stage of MS-TCN tends to be large on high-performance structures. 
In contrast, the average dilation rate in the third stage of MS-TCN is relatively small on high-performance structures.
We assume that the first stage of MS-TCN requires large receptive fields to 
get the long-term context for coarse prediction, while the following stages need small receptive fields to refine the results locally.

\begin{table}[t]
   \small
   \centering 
   \setlength{\tabcolsep}{1.0mm}
   \begin{tabular}{lcccc}  
   \toprule
   & MS-TCN  &\tbf{Arch-50Salads} & \tbf{Arch-GTEA} & \tbf{Arch-BF} \\
   \midrule
   \tbf{50Salads}
   &  67.1  & \tbf{75.4} & 68.8 & 72.6 \\
   \tbf{GTEA}
   & 83.8   & 82.4   & \tbf{88.9} & 85.6 \\
   \tbf{BF}
   & 69.9   & 75.1 & 72.5 & \tbf{76.4} \\
   \bottomrule
   \end{tabular}
   \vspace{2pt}
   \caption{Cross-validation performance (F@0.1) of searched structures among the fold 1 of different datasets. 
      Arch-dataset indicates the structure is searched on which dataset.} 
   \label{tab:ablation_cross_val_datasets}
\end{table}

\begin{table}[t]
   \small
   \centering 
   \setlength{\tabcolsep}{3.3mm}
   \begin{tabular}{lcccc}  \toprule
   \tbf{BreakFast} 
   & Arch-1 & Arch-2 & Arch-3 & Arch-4 \\
   \midrule
   fold1
   & 76.4 & 76.3 & 76.2 & 75.7  \\
   fold2
   & 74.1 & 75.3 & 75.1 & 74.6  \\
   fold3
   & 76.1 & 76.6 & 76.1 & 75.4  \\
   fold4
   & 71.7 & 72.1 & 72.0 & 71.8  \\
   \bottomrule
   \end{tabular}
   \vspace{2pt}
   \caption{Cross-validation performance (F@0.1) of searched structures among different folds of the BreakFast dataset. 
    Arch-n means the structure is searched on fold n.} 
   \label{tab:ablation_cross_val_splits}
\end{table}

\begin{table}[t]
   \small
   \centering 
   \setlength{\tabcolsep}{1.8mm}
   \begin{tabular}{lccccc}  \toprule
   \tbf{50Salads} 
   & F@0.1 & F@0.25 & F@0.5 & Edit & Acc  \\
   \midrule
   \textsl{Spatial CNN~\cite{lea2016segmental}}
   & 32.3 & 27.1 & 18.9 & 24.8 & 54.9 \\
   \textsl{Bi-LSTM~\cite{singh2016multi}}
   & 62.6 & 58.3 & 47.0 & 55.6 & 55.7 \\
   \textsl{Dilated TCN~\cite{lea2017temporal}}
   & 52.2 & 47.6 & 37.4 & 43.1 & 59.3 \\
   \textsl{ST-CNN~\cite{lea2016segmental}}
   & 55.9 & 49.6 & 37.1 & 45.9 & 59.4 \\
   \textsl{TUnet~\cite{ronneberger2015u}}
   & 59.3 & 55.6 & 44.8 & 50.6 & 60.6 \\
   \textsl{ED-TCN~\cite{lea2017temporal}}
   & 68.0 & 63.9 & 52.6 & 59.8 & 64.7 \\
   \textsl{TResNet~\cite{he2016deep}}
   & 69.2 & 65.0 & 54.4 & 60.5 & 66.0 \\
   \textsl{TricorNet~\cite{ding2017tricornet}}
   & 70.1 & 67.2 & 56.6 & 62.8 & 67.5 \\
   \textsl{TRN~\cite{lei2018temporal}}
   & 70.2 & 65.4 & 56.3 & 63.7 & 66.9 \\
   \textsl{TDRN~\cite{lei2018temporal}}
   & 72.9 & 68.5 & 57.2 & 66.0 & 68.1 \\
   \hline
   \textsl{MS-TCN~\cite{farha2019ms}}
   & 76.3 & 74.0 & 64.5 & 67.9 & 80.7 \\
   \textsl{Ours-MS-TCN}
   & \tbf{80.3} & \tbf{78.0} & \tbf{69.8} & \tbf{73.4} & \tbf{82.2} \\
   \bottomrule
   \end{tabular}
   \vspace{2pt}
   \caption{Comparison with SOTA on the 50Salads dataset.} 
   \vspace{-10pt}
   \label{tab:compare_sota_gtea_50}
\end{table}

\section{Conclusion}
We propose a global-to-local search scheme to search for effective receptive
field combinations in a coarse-to-fine scheme.
The global search discovers effective receptive field combinations with better performance than hand designings
but completely different patterns.
The expectation guided iterative local search scheme enables searching 
fine-grained receptive field combinations in the dense search space.
Our proposed global-to-local search can be plugged into multiple tasks, 
\ie action segmentation, probabilistic forecasting~\cite{chen2020probabilistic}, 
classification~\cite{pami20Res2net,gu2021dots,gao2021rbn}, 
segmentation~\cite{GaoEccv20Sal100K,VecRoad_20CVPR,gu2020pyramid}, detection~\cite{ren2015faster,HanEccv20SemLine} methods to further boost the performance.


\vspace{-8pt}
\paragraph{Acknowledgement}

This research was supported by the Major Project for New Generation of AI 
under Grant No. 2018AAA0100400, NSFC (61922046), 
and S\&T innovation project from Chinese Ministry of Education.
We thank MindSpore~\cite{mindspore2020} for the partial support of this work.

\clearpage
{
\small
\bibliographystyle{ieee_fullname}
\bibliography{ref}
}

\end{document}